\pdfoutput=1

\documentclass[11pt]{article}

\usepackage{acl}

\usepackage{times}
\usepackage{latexsym}
\usepackage{booktabs}
\usepackage{graphicx}
\usepackage{pgfplots}
\pgfplotsset{compat=1.17}
\usepackage[T1]{fontenc}

\usepackage[utf8]{inputenc}

\usepackage{microtype}

\usepackage{inconsolata}

\title{MasonTigers at SemEval-2024 Task 9: Solving Puzzles with an Ensemble of Chain-of-Thought Prompts}

\author{Md Nishat Raihan, Dhiman Goswami, Al Nahian Bin Emran, \\ {\bf Sadiya Sayara Chowdhury Puspo, Amrita Ganguly, Marcos Zampieri } \\ George Mason University, USA \\
\texttt{mraihan2@gmu.edu}
}

\begin{document}
\maketitle
\begin{abstract}
This paper presents the \textit{MasonTigers}' submission to the SemEval-2024 Task 9 which provides a dataset of puzzles for testing natural language understanding. We employ large language models (LLMs) to solve this task through several prompting techniques. We show that zero-shot and few-shot prompting with proprietary LLMs outperform open-source models. Results are further improved with chain-of-thought prompting. We obtain our best results by utilizing an ensemble of chain-of thought prompts, ranking $2^{nd}$ in the word puzzle sub-task and $13^{th}$ in the sentence puzzle sub-task. 
\end{abstract}

\section{Introduction}

In recent years, LLMs have achieved impressive performance on several question answering and language understanding tasks when provided with appropriate prompting \cite{brown2020language}. However, complex reasoning abilities often present a challenge for these models. SemEval-2024 Task 9 \cite{jiang-semeval-2024-brainteaser} introduces a novel dataset called \textit{BrainTeaser} \cite{jiang-etal-2023-brainteaser} which includes a set of complex puzzles and brainteasers. Such tasks involve solving word and sentence puzzles, which require multi-step inference and deduction. The dataset covers a diverse range of puzzle types including sequences, analogies, classification, mathematical reasoning, inferences about implicit relationships, and more. Solutions frequently demand a chained application of knowledge and logic across multiple steps to uncover insights or concepts not directly stated in the problem description. 

Solving these elaborate reasoning problems is a challenging scenario for NLP systems. We explore whether and how LLMs can succeed on this task. We employ proprietary models such as GPT-4 \cite{openai2023gpt4} and Claude 2.1 \cite{anthropic2023claude} through APIs. These models have shown promising few-shot reasoning ability. We also use Mixtral \cite{jiang2024mixtral}, an open-source LLM that shows state-of-the-results in several language reasoning tasks. The prompting paradigm involves providing models with natural language descriptions that encode the reasoning process step-by-step \cite{liu2021makes}. We implement various prompting approaches for mapping puzzles to conditional text and systematically transforming reasoning into explanation chains. Our core method, chain-of-thought prompting \cite{wei2022chain}, iteratively breaks down the deduction into simplified logical steps.

Experiments reveal that while zero-shot performance lags due to a lack of grounding, multi-step prompts can unlock substantial reasoning ability in models. Performance improves with more steps and more specificity in the prompt. While introducing few-shot prompting generates good results, we observed that models do significantly better with chain-of-thought prompting. We experiment with several chains of thought and achieve mostly similar results with each attempt. To make a more empirically confident guess towards solving the puzzles we adopt an ensemble of these chains based on majority voting. Our approach achieves competitive performance, ranking \textit{2nd} on the word puzzle subtask and \textit{13th} on sentence puzzles.

\section{Related Work}

LLMs have been widely used for complex and challenging language processing tasks recently \cite{raihan2023sentmix, raihan2023offensive, goswami2023offmix}. They have shown good reasoning abilities in several tasks. The task of solving puzzles and the BrainTeaser dataset \cite{jiang-etal-2023-brainteaser} represent both a novel task and a novel dataset respectively. Similarly to their multiple choice questions (MCQs) approach, a few datasets like MathQA \cite{austin2021program}, have been compiled. However, these are intended for specific tasks in which domain knowledge is usually enough thus they not requiring deep reasoning. A similar work is done by \citet{saeedi2020csnlp} where they investigate a task that combines natural language understanding and commonsense reasoning. They present deep learning architectures for distinguishing between sensible and nonsensical statements.

Pun detection by \cite{zou-lu-2019-joint} is a puzzle-like activity that is similar to BrainTeaser. It presents a method for joint pun detection and localization utilizing a sequence labeling perspective. This highlights the complexity of language comprehension, especially in detecting subtle wordplay. Another dataset, LatEval is curated by \citet{huang2023lateval} that delves further into lateral thinking and commonsense reasoning, highlighting the challenges faced by language models in tasks requiring unconventional thinking and creativity. \citet{zhou2023rome} presents ROME, a dataset designed to assess vision-language models' capacity to reason beyond intuitive understanding, highlighting the shortcomings of existing models in understanding events that defy common sense.

In the field of reasoning task, a \textit{chain-of-thought} \cite{wei2022chain} implies a logical sequence of connected ideas, fostering coherence and depth in responses. On the other hand, a \textit{tree-of-thought} suggests branching out into various related ideas, offering a more comprehensive exploration of a topic. While few-shot prompting is effective for some tasks by providing examples to guide the model, it may have limitations in capturing the complexity of nuanced conversations. The optimal choice may involve a hybrid approach, where a few-shot prompt sets the initial \cite{yao2023tree} context, and the model subsequently follows a chain or tree of thought to generate more contextually rich and coherent responses useful for reasoning tasks.

\citet{tan2023causal} shows the performance of LLM's on the reasoning of arithmetic word problems. It states that higher degrees of realization are associated with better overall accuracy on arithmetic problems. And chain-of-thought is really helpful in this aspect as it covers a variety of prompts to strengthen the reasoning. Similarly, \citet{mo2023tree} presents a new reasoning framework for large language models by addressing a gap in prior tree-based reasoning methods which overlooked inherent uncertainties in intermediate decision points made by models. Overall, the key innovation is leveraging uncertainty estimation locally within the models during tree reasoning to enable more precise problem-solving and reasoning.

\section{The BrainTeaser Dataset}

The BrainTeaser dataset \cite{jiang-etal-2023-brainteaser}, introduced with the task \cite{jiang-semeval-2024-brainteaser} is a question-answering benchmark designed to evaluate models' ability for lateral thinking, i.e., to defy default commonsense associations and reason creatively. The dataset contains 1,100 multiple-choice questions divided into two sub-tasks - 627 sentence-based puzzles relying on narrative context and common phrases and 492 puzzles focused on the literal form and letters of words.

For a fair comparison with human performance, the dataset also provides a separate human evaluation set with 102 randomly sampled questions. Each question in BrainTeaser has one correct answer and three distractor choices, including the option "none of the above". To prevent memorization of training data, the dataset also contains semantically and contextually reconstructed variants for every question while preserving the original reasoning process and answers. The key statistics of the dataset are shown in Table \ref{tab:stats}.

\begin{table}[!ht]
\begin{center}
\begin{tabular}{@{}lcc@{}}
\toprule
& \bf Sentence & \bf Word \\
\midrule
Number of puzzles & 627 & 492 \\
Avg. tokens (prompt) & 34.88 & 10.65 \\
Avg. tokens (choices) & 9.11 & 3.0 \\
\bottomrule
\end{tabular}
\end{center}
\caption{Key statistics of the BrainTeaser dataset in the sentence and word puzzle sub-task.}
\label{tab:stats}
\end{table}

\noindent During the SemEval-2024 Task 9 development phase, a total of 240 prompts (120 for both sentence and word puzzles) are provided. During the test phase, a total of 216 prompts (120 for sentence and 96 for word puzzles) are provided.

\section{Experiments}

In our experiments, we focus on several prompting strategies by employing three state-of-the-art models including proprietary models like GPT-4 \cite{openai2023gpt4} and Claude 2.1 \cite{anthropic2023claude} (accessed via API key) and one open-source model - Mixtral \cite{jiang2024mixtral}. 

\subsection{Zero-Shot Prompting}

We start with zero-shot prompting by assigning the AI a role, describing the task, and giving it one puzzle at a time, as shown in Figure \ref{fig:prompt1}.

\begin{figure}[h]
\centering
\scalebox{.92}{
\begin{tikzpicture}[node distance=1cm]
    \tikzstyle{block} = [rectangle, draw, fill=blue!10, text width=\linewidth, text centered, rounded corners, minimum height=4em]
    \tikzstyle{operation} = [text centered, minimum height=1em]
    \node [block] (rect1) {\textbf{Role:}\\{ You are a helpful AI assistant. You are given the task of solving a sentence/word puzzle. }};
    \node [operation, below of=rect1] (plus1) {+};
    \node [block, below of=plus1] (rect2) {\textbf{Definition:}\\{ It's a Multiple Choice Question paired with 4 potential answers. Choose the most suitable one. }};
    \node [operation, below of=rect2] (plus3) {+};
    \node [operation, below of=plus3] (plus4) {};
    \node [block, below of=plus4] (rect3) {{\textbf{Question:} What part of London is in France? \\ \textbf{Choices:} \\ 1. The letter N. \\ 2. The letter O. \\ 3. The letter L. \\ 4. None of the above. \\ Choose the most suitable answer. Thanks.}};
\end{tikzpicture}
}
\caption{Sample structure for Zero-Shot Prompting.}
\label{fig:prompt1}
\end{figure}

\subsection{Few-Shot Prompting}
In order to give the LLMs more context we integrate more examples and design prompts for few-shot prompting. We include 4 solved puzzles from the train set and then attach one puzzle from the test set each time we prompt the models. We also use some tags for better extracting the generated answers, as shown in Figure \ref{fig:prompt2}.

\begin{figure}[h]
\centering
\scalebox{.92}{
\begin{tikzpicture}[node distance=1cm]
    \tikzstyle{block} = [rectangle, draw, fill=blue!10, text width=\linewidth, text centered, rounded corners, minimum height=4em]
    \tikzstyle{block1} = [rectangle, draw, fill=green!60!blue!10!, text width=\linewidth, text centered, rounded corners, minimum height=4em]
    \tikzstyle{operation} = [text centered, minimum height=1em]
    \node [block] (rect1) {\textbf{Role:} \\ { You are a helpful AI assistant. You are given the task of solving a sentence/word puzzle. }};
    \node [operation, below of=rect1] (plus1) {+};
    \node [block, below of=plus1] (rect2) {\textbf{Definition:} \\ { It's a Multiple Choice Question paired with 4 potential answers. Choose the most suitable one. }};
    \node [operation, below of=rect2] (plus3) {+};
    \node [operation, below of=plus3] (plus4) {};
    \node [operation, below of=plus4] (plus5) {};
    \node [operation, below of=plus5] (plus6) {};
    \node [block1, below of=plus6] (rect3) { \textbf{Examples:} \\ {\textbf{Question1:} A man preserves a lengthy beard despite shaving every day. \\	 1. He is a barber. \\ 2. He wants to maintain his appearance. \\ 3. He wants his girlfriend to buy him a razor. \\ 4. None of the above. \\ \textbf{Correct Answer:} \textbf{<ans1>} 1 \textbf{</ans1>} \\ ..... \\ \textbf{Question4:} Tom attends class every day but doesn’t do any homework. \\	1. He is a teacher. \\ 2. He is a lazy person. \\ 3. His teacher will not let him fail. \\ 4. None of the above. \\ \textbf{Correct Answer:} \textbf{<ans4>} 1 \textbf{</ans4>}}};
    \node [operation, below of=rect3] (plus7) {};
    \node [operation, below of=plus7] (plus8) { };
    \node [operation, below of=plus8] (plus9) { };
    \node [operation, below of=plus9] (plus10) { + };
    \node [operation, below of=plus10] (plus12) { };
    \node [operation, below of=plus12] (plus13) {+};
    \node [block, below of=plus13] (rect4) {{\textbf{Question5:} The brother of a beggar passed away, but the deceased had no brothers. How is that possible? \\ \textbf{Choices:} \\ 1. The beggar was the man's sister. \\ 2. The man is angry for his brother being a beggar and cut ties with him. \\ 3. The bagger's brother is a murderer. \\ 4. None of the above. \\ \textbf{Correct Answer:} \textbf{<ans5>} ? \textbf{</ans5>} Choose the most suitable answer. Thanks. \\}};
\end{tikzpicture}
}
\caption{Sample structure for Few-Shot Prompting.}
\label{fig:prompt2}
\end{figure}

\subsection{Chain-of-Thought}
\label{cot}
To guide the models toward better reasoning - we experiment with chain-of-thought prompting. We give the model the puzzle, and the potential answers and work with every example one-by-one in order to choose the most reasonable one. Like the original CoT approach \cite{wei2022chain}, we do not assign any role or explain the task - just pose the question, the CoT, and the answer (see Figure \ref{fig:prompt3}). We do this as 2-shot, 4-shot, and 8-shot for all three models.

\begin{figure}[!h]
\centering
\scalebox{.92}{
\begin{tikzpicture}[node distance=5.2cm]
    \tikzstyle{block} = [rectangle, draw, fill=green!60!blue!10!, text width=\linewidth, text centered, rounded corners, minimum height=4em]
    \tikzstyle{operation} = [text centered, minimum height=1em]
    \node [block] (rect1) {{\textbf{\underline{Question 1:}} How do you spell COW in thirteen letters? \\ -- \\ \textbf{\underline{Choices 1:}} \\ 1. SEE OH DEREFORD. \\ 2. SEE O DOUBLE YOU. \\ 3. COWCOWCOWCOWW. \\ 4. None of above. \\ -- \\ \textbf{\textit{\underline{Chain-of-Thought 1:}}} \\  \underline{1. SEE OH DEREFORD:} Doesn't seem to spell out "COW" in any conventional or playful manner. \\ \underline{2. SEE O DOUBLE YOU:} Spells out "COW" in a creative way, matching the letter count required. \\ \underline{3. COWCOWCOWCOWW:} Simply repeats the word "COW" without cleverly meeting the thirteen-letter criteria. \\ \underline{4. None of the above:} Not applicable since there is a viable option. \\ -- \\ \underline{\textbf{Decision 1:}} The answer "SEE O DOUBLE YOU" creatively spells "COW" using thirteen letters, making it the correct choice. \\ -- \\ \underline{\textbf{Answer 1:}} 2. \\ ..... \\ ..... \\ ..... \\ ..... \\ \textbf{\underline{Question 8:}} -- \\ \textbf{\underline{Choices 8:}} -- \\ \textbf{\textit{\underline{Chain-of-Thought 8:}}} -- \\  \textbf{\underline{Decision 8:}} -- \\ \textbf{\underline{Answer 8:}} -- \\ ------ \\ ------ \\ \textbf{\underline{Question 9:}} How do you spell COB in seven letters? \\ -- \\ \textbf{\underline{Choices 9:}} \\ 1. COBCOBB \\ 2. COBBLER \\ 3. SEE O BEE. \\ 4. None of the above. \\ --  }};
\end{tikzpicture}
}
\caption{Sample structure for Chain-of-Thought Prompting (8-shot).}
\label{fig:prompt3}
\end{figure}

\subsection{Ensemble of Chain-of-Thought Prompts}
\label{ens}


To assess model performance, an ensemble approach is utilized with chain-of-thought prompting to make more confident guesses regarding the correct answers. Specifically, majority voting is done across an ensemble of models prompted by different question groups. For each prompt, 8 different random questions are selected from the BrainTeaser training set - repeated 5 times in total. Finally, the predictions are aggregated through voting to output the overall ensemble prediction.

This ensemble methodology with chain-of-thought prompting helps improve robustness to outlier examples and noise compared to using a single model. By prompting the ensemble components on different random question subsets, diversity is promoted to capture a more holistic representation of the overall data distribution. The voting also helps cancel out issues with single models latching onto spurious patterns. Overall, the ensemble approach with multiple chain-of-thought prompt sets provides a robust assessment strategy suited for the open-ended nature and diversity of the BrainTeaser puzzles.


\section{Results}

We analyze the performance of the three models - including GPT4 Turbo, Claude 2.1, and Mixtral. These models are tested with 
different types of prompts - regular and chain-of-thought, and with a varying number of examples, known as shots, ranging from zero to eight. Additionally, an ensemble method is applied to the eight-shot chain-of-thought prompting to see if it can further improve the models' performance. The results, shown in Table \ref{tab:res}, reveal how the models performed under each condition. A human baseline with scores of 0.91 for both Sentence and Word puzzles in the test set is provided by the task organizers for comparison purposes.

\begin{table*}[!h]
\centering
\scalebox{0.9}{%
\begin{tabular}{lcc|cc|cc}
\toprule
\textbf{Model} & \textbf{Prompting} & \# of Shot & \textbf{Sen\_Dev} & \textbf{Sen\_Test} & \textbf{Word\_Dev} & \textbf{Word\_Test} \\
\midrule
Human Baseline & -- & -- & -- & 0.91 & -- & 0.91\\
\midrule
GPT4 Turbo & Regular & Zero Shot & 0.79 & 0.76 & 0.81 & 0.79\\
GPT4 Turbo & Regular & 4 Shot & 0.90 & 0.91 & 0.87 & 0.86\\
GPT4 Turbo & CoT & 2 Shot & 0.87 & 0.88 & 0.85 & 0.89\\
GPT4 Turbo & CoT & 4 Shot & 0.89 & 0.90 & 0.92 & 0.91\\
GPT4 Turbo & CoT & 8 Shot & 0.93 & 0.92 & 0.94 & 0.94\\
GPT4 Turbo & CoT \textbf{[E]} & 8 Shot & \textbf{0.94} & \textbf{0.93} & \textbf{0.96} & \textbf{0.95} \\
\midrule
Claude 2.1 & Regular & Zero Shot & 0.76 & 0.77 & 0.71 & 0.62\\
Claude 2.1 & Regular & 4 Shot & 0.87 & 0.84 & 0.87 & 0.85\\
Claude 2.1 & CoT & 2 Shot & 0.84 & 0.81 & 0.83 & 0.84\\
Claude 2.1 & CoT & 4 Shot & 0.91 & 0.84 & 0.90 & 0.94\\
Claude 2.1 & CoT & 8 Shot & 0.90 & 0.84 & 0.90 & 0.94\\
Claude 2.1 & CoT \textbf{[E]} \textbf{[*]} & 8 Shot & \textit{\textbf{0.91}} & \textit{\textbf{0.86}} & \textit{\textbf{0.91}} & \textit{\textbf{0.95}} \\
\midrule
Mixtral & Regular & Zero Shot & 0.71 & 0.66 & 0.45 & 0.51\\
Mixtral & Regular & 4 Shot & 0.81 & 0.82 & 0.79 & 0.75\\
Mixtral & CoT & 2 Shot & 0.79 & 0.75 & 0.63 & 0.70\\
Mixtral & CoT & 4 Shot & 0.84 & 0.86 & 0.77 & 0.76\\
Mixtral & CoT & 8 Shot & 0.89 & 0.86 & 0.80 & 0.81\\
Mixtral & CoT \textbf{[E]} & 8 Shot & 0.89 & 0.88 & 0.81 & 0.82\\
\bottomrule
\end{tabular}%
}
\caption{Comparing the results generated by the models with different prompting strategies. \textbf{[CoT]} - denotes chain-of-thought. \textbf{[E]} - denotes Ensemble (as described in \ref{ens}). \textbf{[*]} - denotes submission during the test phase on the Leaderboard. }
\label{tab:res}
\end{table*}

GPT4 Turbo shows the best performance, especially with chain-of-thought prompting and an increasing number of shots. The model performs best with the eight-shot chain-of-thought prompting combined with the ensemble method ([E]), reaching the highest Sentence and Word scores of 0.93 and 0.95 in the test set, respectively. This shows that chain-of-thought prompting and the ensemble method significantly improve the model's understanding and output. Claude 2.1 also improves with chain-of-thought prompting and more shots. Its best scores were with the eight-shot chain-of-thought with the ensemble, achieving Sentence and Word scores of 0.86 and 0.95 in the test set, respectively. The asterisk (*) mark in Table \ref{tab:res} denotes our submission during the test phase. Even though Mixtral's performance is inferior to the performance of the other two models, it consistently gets better with more shots and chain-of-thought prompting.  Mixtral delivered best results with the eight-shot chain-of-thought and the ensemble technique, with Sentence and Word scores of 0.88 and 0.82 in the test set, respectively. 

Finally, the results highlight the effectiveness of chain-of-thought prompting in boosting the performance of LLMs. This approach, especially when combined with more examples and the ensemble method, greatly improves models' abilities to process and generate more accurate responses. GPT4 Turbo's top performance is likely due to its advanced design, which makes the most of these strategies. On the other hand, Claude 2.1's results point to the importance of model-specific adjustments. 

\section{Conclusion and Future Work}

In this paper, we presented MasonTigers' approach to SemEval-2024 Task 9 on solving puzzles using LLMs. We explored various prompting strategies to guide the models, including zero-shot, few-shot, and chain-of-thought prompting. Our key method involved iteratively breaking down reasoning into simplified logical steps to decompose the complex deduction process.

Our experiments revealed promising results. While zero-shot performance was limited, providing explanatory prompts substantially improved the models' reasoning abilities. Performance increased with more prompt specificity and steps. Our best results came from an ensemble approach applying majority voting across multiple chain-of-thought prompts.

Ultimately, our system achieved competitive rankings on the leaderboard, placing $2^{nd}$ in the word puzzle sub-task and $13^{th}$ on sentence puzzles. The strong capability unlocked through guided prompting highlights these models' latent reasoning potential when given a structured thought process. Our work sheds light on how explanatory chains can elicit more of the knowledge within large language model parameters.

A few key limitations remain to be addressed in future work. First, constructing effective prompts requires extensive human effort and insight - automating this prompting process could improve scalability. Additionally, performance still lags behind human levels, indicating that there is room for advancement. Architectural constraints related to long-term memory and reasoning likely need to be overcome. Finally, our approach focused narrowly on the given puzzles rather than teaching broader inferential skills - developing more generalizable reasoning through prompts is an open challenge. 

\section*{Acknowledgments}

We would like to thank the shared task organizers for proposing this interesting competition and for providing participants with the BrainTeaser dataset. 



\bibliography{custom}

\end{document}